\pdfoutput=1

\documentclass[11pt]{article}

\newcommand{\easyedit}[0]{\textsc{EasyEdit}}

\usepackage[]{EMNLP2023}

\usepackage{times}
\usepackage{latexsym}
\usepackage{amssymb}

\usepackage{makecell}
\usepackage{fontawesome}
\usepackage{booktabs}
\usepackage{bbm}
\usepackage{graphicx}
\usepackage{multirow}
\usepackage{caption}
\usepackage[T1]{fontenc}

\usepackage[utf8]{inputenc}

\usepackage{microtype}
\usepackage{amsmath}
\usepackage{inconsolata}
\usepackage{colortbl}
\usepackage{xcolor}

\usepackage{times}
\usepackage{graphicx}
\usepackage{multirow}
\usepackage{latexsym}
\definecolor{easyedit}{RGB}{47, 85, 151}

\newcommand*{\img}[1]{%
    \raisebox{-.5\baselineskip}{%
        \includegraphics[
        height=34pt,
        width=44pt,
        ]{#1}
    }
}

\title{\img{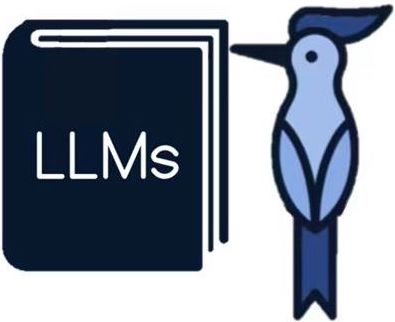}$\hspace{-0.05in}\mathbf{\textcolor{easyedit}{EasyEdit}}$: An Easy-to-use Knowledge Editing Framework for Large Language Models}

\author{
Peng Wang{$^\clubsuit$}, Ningyu Zhang{$^\clubsuit \footnotemark[1]$}, {\bf Bozhong Tian}{$^\clubsuit$}, {\bf Zekun Xi}{$^\clubsuit$}, Yunzhi Yao{$^\clubsuit$}, \\
{\bf Ziwen Xu}{$^\clubsuit$}, {\bf Mengru Wang}{$^\clubsuit$}, {\bf Shengyu Mao}{$^\clubsuit$}, {\bf Xiaohan Wang}{$^\clubsuit$}, 
{\bf Siyuan Cheng}{$^\clubsuit$},\\ {\bf Kangwei Liu}{$^\clubsuit$}, {\bf Yuansheng Ni}{$^\clubsuit$},  {\bf Guozhou Zheng}{$^\clubsuit$}, {\bf Huajun Chen}$^{\clubsuit}\thanks{~~Corresponding author.}$, \\
 $^\clubsuit$ Zhejiang University \\
  \faGithub \, \normalsize{\url{https://github.com/zjunlp/EasyEdit}}
  }

\begin{document}
\maketitle
\begin{abstract}
Large Language Models (LLMs) usually suffer from knowledge cutoff or fallacy issues, which means they are unaware of unseen events or generate text with incorrect facts owing to outdated/noisy data. To this end, many knowledge editing approaches for LLMs have emerged -- aiming to subtly inject/edit updated knowledge or adjust undesired behavior while minimizing the impact on unrelated inputs. Nevertheless, due to significant differences among various knowledge editing methods and the variations in task setups, there is no standard implementation framework available for the community, which hinders practitioners from applying knowledge editing to applications. To address these issues, we propose \easyedit, an easy-to-use knowledge editing framework for LLMs. It supports various cutting-edge knowledge editing approaches and can be readily applied to many well-known LLMs such as T5, GPT-J, LlaMA, etc. Empirically, we report the knowledge editing results on LlaMA-2 with \easyedit, demonstrating that knowledge editing surpasses traditional fine-tuning in terms of reliability and generalization. We have released the source code on GitHub\footnote{This is a subproject of KnowLM (\url{https://github.com/zjunlp/KnowLM}), which facilitates knowledgeable LLM Framework with EasyInstruct, EasyEdit, EasyDetect etc.}, along with Google Colab tutorials and comprehensive documentation\footnote{\url{https://zjunlp.gitbook.io/easyedit}} for beginners to get started. Besides, we present an online system\footnote{\url{https://huggingface.co/spaces/zjunlp/EasyEdit}} for real-time knowledge editing, and a demo video\footnote{\url{https://youtu.be/Gm6T0QaaskU}}.
\end{abstract}

\section{Introduction}

Large Language Models (LLMs) have revolutionized modern Natural Language Processing (NLP), significantly improving performance across various tasks \cite{brown2020language, openai2023gpt4, anil2023palm, zhao2023survey, touvron2023llama2, qiao-etal-2023-reasoning,DBLP:journals/corr/abs-2307-04964,DBLP:journals/corr/abs-2306-08302}.
However, deployed LLMs usually suffer from knowledge cutoff or fallacy issues.
For example, LLMs such as ChatGPT and LlaMA possess information only up to their last training point. 
They can sometimes produce inaccurate or misleading information due to potential discrepancies and biases in their pre-training data \cite{Ji_2023,hartvigsen-etal-2022-toxigen}.
Hence, it's essential to efficiently update the parametric knowledge within the LLMs to modify specific behaviors while avoiding expensive retraining.


\begin{figure*}[t]
    \centering
    \includegraphics[width=1\textwidth]{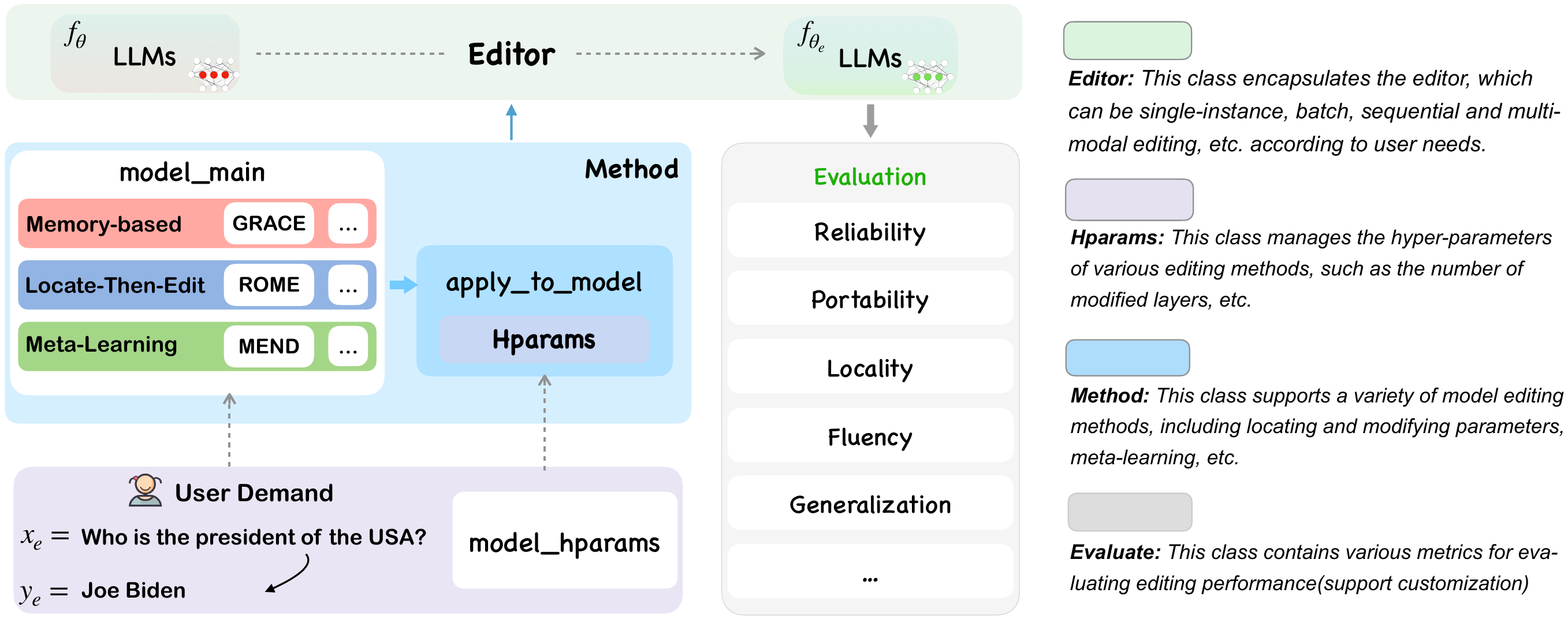}
    \caption{The overall architecture of \easyedit.
    The main function is \texttt{apply\_to\_model}, which applies the selected editing method to the LLMs. 
    The \textbf{Editor} serves as the direct entry point, receiving customized user inputs and outputs, and returning the edited weights. 
    Please note that some methods may require pre-training of classifiers or hypernetworks through the Trainer (See \S{\ref{sec:trainer}}). \easyedit ~supports customizable evaluation metrics.}
    \label{fig:framework}
\end{figure*}

Indeed, finetuning or parameter-efficient finetuning \cite{ding2022delta, ding2023parameter} offers methods for modifying LLMs, these approaches can be computationally expensive and may lead to overfitting, particularly when applied to a limited number of samples \cite{decao2021editing} or streaming errors of LLMs.
Additionally, fine-tuned models might forfeit capabilities gained during pre-training, and their modifications do not always generalize to relevant inputs.
An alternative methodology involves using manually written or retrieved prompts to influence the LLMs' output.
These methods suffer from reliability issues, as LLMs do not consistently generate text aligned with the prefix prompt \cite{hernandez2023inspecting, lewis2021retrievalaugmented}.
Additionally, due to the extensive amount of up-to-date knowledge required for complex reasoning tasks, the impracticality of context overload becomes inevitable whenever the context length is limited.

A feasible solution, knowledge editing\footnote{Knowledge editing can also be termed as model editing.}, aims to efficiently modify the behavior of LLMs with minimal impact on unrelated inputs.
Research on knowledge editing for LLMs \cite{meng2023locating, meng2022massediting, zheng2023edit, DBLP:journals/corr/abs-2305-14956, mitchell2022fast,DBLP:journals/corr/abs-2304-14767,DBLP:journals/corr/abs-2301-04213,DBLP:journals/corr/abs-2307-12976, hartvigsen2023aging, tan2024massive, yu2023melo} have displayed remarkable progress across various tasks and settings. 

However, these variations in both implementation and task settings have impeded the development of a unified and comprehensive framework for knowledge editing.
Note that the complexity obstructs the direct comparison of effectiveness and feasibility between different methods, and complicates the creation of novel knowledge editing approaches.
To this end, we propose \easyedit, an easy-to-use knowledge editing framework for LLMs. 
\easyedit ~modularizes editing methods and effectiveness evaluation while considering their combination and interaction.
It supports a variety of editing scenarios, including \textbf{single-instance}, \textbf{batch-instance}, \textbf{sequential}, and \textbf{multi-modal} editing.
Moreover, \easyedit ~provides evaluation evaluations of key metrics such as Reliability, Generalization, Locality, and Portability \cite{yao2023editing}, to quantify the robustness and side effects \cite{cohen2023evaluating} of editing methods.

Specifically, in \easyedit, the Editor class integrates various editing components.
The Method class offers a unified interface \texttt{apply\_to\_model}, which accepts editing descriptors and returns the edited model, thereby facilitating the integration of novel editing methodologies.
Dedicated to evaluating editing performance, the Evaluate module leverages metrics such as reliability, robust generalization, and locality.
The Trainer module manages the training of additional neural network structures.
Each module in \easyedit ~is meticulously defined, striking a balance between cohesion and coupling. 
Furthermore, we furnish examples of editing across a spectrum of models, including T5 \cite{DBLP:journals/corr/abs-1910-10683}, GPT-J \cite{gpt-j}, GPT-NEO \cite{gpt-neo}, GPT2 \cite{radford2019language}, LLaMA \cite{touvron2023llama1}, LLaMA-2 \cite{touvron2023llama2}, Mistral \cite{jiang2023mistral}, and Qwen \cite{bai2023qwen}.
We acknowledge all the support for {\easyedit}, which is listed in Appendix \ref{sec:ack} due to space constraints.

\section{Background}

\paragraph{Previous Solutions}
Despite the tremendous success of LLMs in almost all NLP tasks, persistent challenges such as knowledge cutoff and biased/toxic outputs remain.
To counter these challenges, two approaches are generally employed:

1) \textsc{Fine-tuning}: Traditional fine-tuning techniques, along with delta tuning \cite{ding2022delta} and LoRA tuning \cite{hu2021lora} utilize domain-specific datasets to update the model's internal parametric knowledge.
However, these methods face two notable challenges: First, they consume considerable resources.
Second, they risk the potential of catastrophic forgetting \cite{ramasesh2022effect}.

2) \textsc{Prompt-Augmentation}: Given a sufficient number of demonstrations or retrieved contexts, LLMs can learn to enhance reasoning \cite{DBLP:conf/emnlp/0002ZZW0F022} and generation through external knowledge \cite{borgeaud2022improving, guu2020retrieval, lewis2020retrieval}.
However, the performance may be sensitive to factors such as the prompting template, the selection of in-context examples \cite{pmlr-v139-zhao21c}, or retrieved contexts \cite{ren2023investigating}.
These approaches also encounter the issue of context length limitation \cite{liu2023lost}.

\paragraph{Knowledge Storage Mechanism}
Within the NLP literature, numerous studies have delved into understanding the location of different types of knowledge in language models \cite{petroni-etal-2019-language, roberts-etal-2020-much, jiang-etal-2020-know}. 
LLMs can be conceptualized as knowledge banks, and the transformer MLP layers function as key-value memories according to observations from \citet{geva-etal-2021-transformer}.
This configuration promotes efficient knowledge adjustments by precisely localizing knowledge within the MLP layers (denoted as knowledge editing).

Knowledge editing enables nimble alterations to the LLMs' behavior through one data point. Another promising attribute of knowledge editing is its ability to ensure the locality of editing, meaning that modifications are contained within specific contexts. Additionally, the knowledge editing technique can mitigate harmful language generation \cite{geva-etal-2022-transformer}. 
In this paper, we present \easyedit, an easy-to-use knowledge editing framework for LLMs.
It seamlessly integrates diverse editing technologies and supports the free combination of modules for various LLMs.
Through its unified framework and interface, \easyedit ~enables users to swiftly comprehend and apply the prevalent knowledge editing methods included in the package.

\section{Design and Implementation}
\easyedit ~provides a complete editing and evaluation process built on Pytorch \cite{paszke2019pytorch} and Huggingface \cite{wolf2020huggingfaces}.
This section commences with an exploration of the assemblability aspect of \easyedit, followed by a detailed explanation of the design and implementation of each component within the \easyedit ~framework (as shown in Figure \ref{fig:framework}).
Additionally, we demonstrate a straightforward example of applying MEND to LLaMA, altering the output of \emph{the U.S. President} to \emph{Joe Biden}.

\subsection{Assemblability}

\begin{figure}[t]
    \centering
    \includegraphics[width=0.5\textwidth]{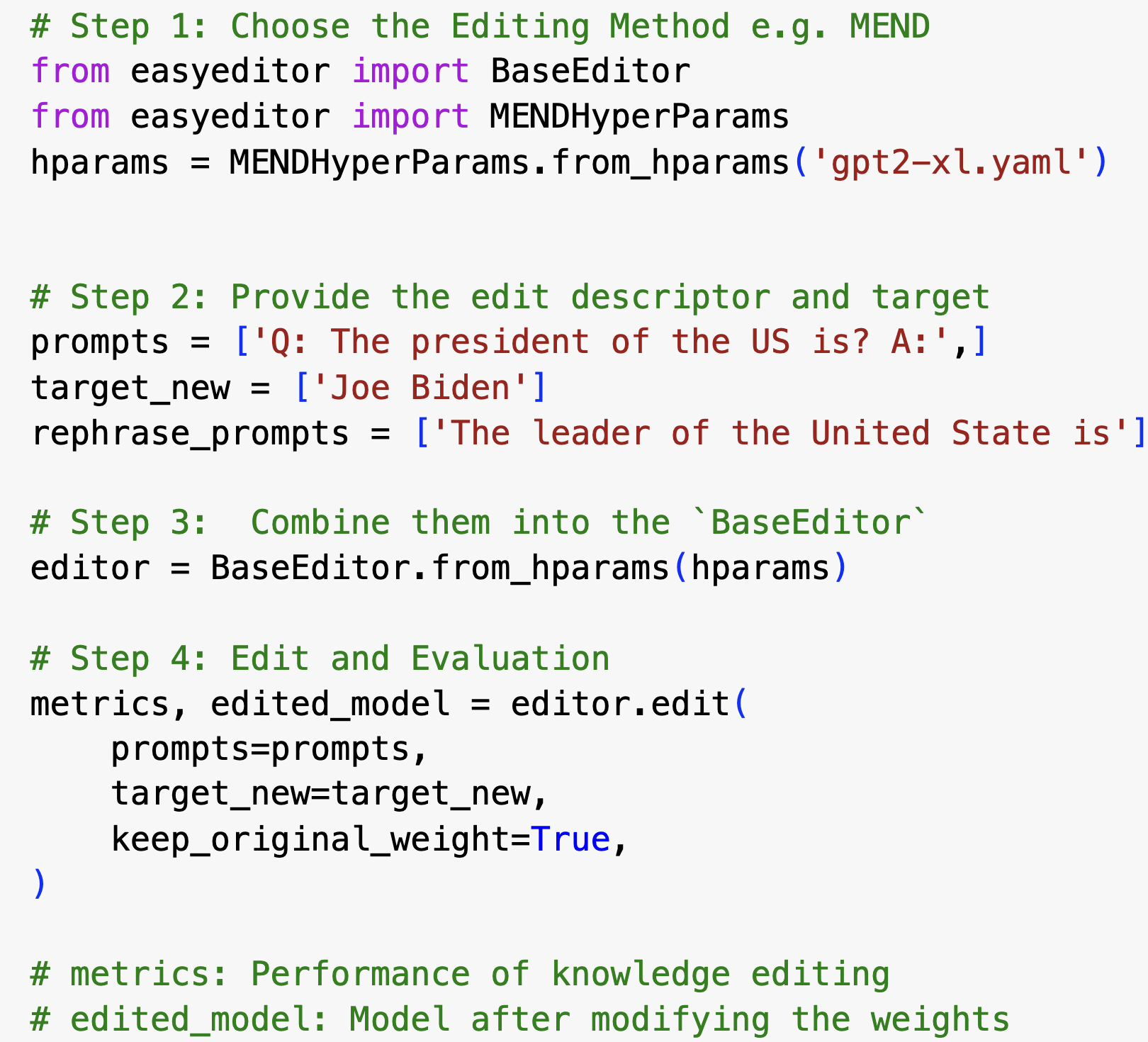}
    \caption{
    A running example of knowledge editing for LLMs in \easyedit. 
    Utilizing the MEND approach, we can successfully transform the depiction of \emph{the U.S. President} into that of \emph{Joe Biden}.}
    \vspace{-0.15in}
    \label{fig:code_example}
\end{figure}

In the realm of knowledge editing, various distinct scenarios\footnote{Denoted as \texttt{(Editor, METHOD, TARGET)}} exist.
To cater to this diversity, \easyedit ~offers flexible combinations of modules that different editing \texttt{Editor} (such as single-instance, batch-instance (details in Appendix \ref{sec:editing_definition})), \texttt{METHOD} (such as ROME, GRACE (\S{\ref{sec:method}})).
About editing \texttt{TARGET}, \easyedit ~can accommodate any parameterized white-box existing model. 
Additionally, recent research \cite{dong2022survey} indicates that LLMs exhibit robust in-context learning capabilities.
By providing edited facts to LLMs, one can alter the behavior of black-box models such as GPT4 \cite{openai2023gpt4}.
All those combinations are easily implementable and verifiable within the \easyedit ~framework.

\begin{table*}[!tp]
\centering
\resizebox{\textwidth}{!}{
\begin{tabular}{llcccccc}
\toprule
&\textbf{Method}  & \textbf{\makecell{Batch\\Edit}} & \textbf{\makecell{Sequential\\Edit}} & \textbf{\makecell{Additional\\Train}} & \textbf{\makecell{Edit\\Area}} & \textbf{Time (s)} & \textbf{VRAM (GB)}   \\ \hline
\multirow{4}{*}{Memory-based}
& SERAC  & YES & YES & YES        & \emph{External Model}    & 8.46 & 42  \\ 
&  IKE & NO & NO  & YES & \emph{In-Context}    & 4.57 & 52   \\
&  GRACE & NO & YES  & NO & \emph{MLP+codebook}    & 142.68 & 28   \\
&  MELO & YES & YES  & NO & \emph{LoRA+codebook}    & 154.32 & 30   \\
 \hline
\multirow{2}{*}{Meta-learning}  
& KE       &    YES   &    YES & YES    & \emph{MLP}    & 7.87 & 49    \\
&  MEND     & YES  & YES & YES & \emph{MLP}   & 6.39 & 46    \\
 \hline
\multirow{4}{*}{Locate-Then-Edit}  
& KN & NO  & YES  & NO      & \emph{MLP}   & 425.64 & 42     \\
& ROME    & NO  & YES & NO & \emph{MLP}   & 187.90 & 31      \\
&  MEMIT   & YES  & YES & NO & \emph{MLP}   & 169.28 & 33    \\
&  PMET   & YES  & YES & NO & \emph{MLP}   & 219.17 & 34    \\ 
\bottomrule
\end{tabular}
}
\caption{
Comparison of several model editing methods. 
`Batch Edit' refers to simultaneously editing multiple target knowledge instances. `Sequential Edit' refers to maintaining previously edited knowledge while performing new edits.
`Additional Train' refers to the need for pre-training other network structures or parameters before editing. `Edit Area' indicates the location of the edit, with MLP representing the linear layer.
`Time \& VRAM' reflects the efficiency of the editing method (using LlaMA-7B as an example). `Time' indicates the wall clock time required for conducting 10 edits, while VRAM represents the graphics memory usage.
}
\label{tab:edit_classifier}
\end{table*}

\subsection{Editor}
The ~\texttt{Editor} serves a pivotal role in knowledge editing as it directly establishes the editing tasks and corresponding editing scenarios.
Users supply the editor descriptor ($x_e$) and the edit target ($y_e$), but the input format varies according to the different editing objects.
For instance, in Seq2Seq models, the edit target typically serves as the decoder's input, while in autoregressive models, $x_e$ and $y_e$ need to be concatenated to maximize the conditional probability.
To facilitate unified editing across diverse architecture models, we meticulously develop a component \texttt{prepare\_requests} to transform editing inputs.

In \easyedit, we provide an ``edit'' interface, incorporating components such as \texttt{Hparams}, \texttt{Method}, and \texttt{Evaluate}.
During the editing phase, various knowledge editing strategies can be executed by invoking the \texttt{apply\_to\_model } function available in all different methods, it also performs evaluations of the model before and after the editing to gauge the editing's multifaceted impact on the model behavior, including generalization and side effects. 
An example to edit through \easyedit ~is depicted in Figure \ref{fig:code_example}.

Note that the ability to execute batch editing (multiple edits in a single instance) and sequential editing (implementing new edits while preserving previous editing) is a crucial feature of knowledge editing \cite{huang2023transformerpatcher}.
For methods that support batch editing, editing instances are inputted in chunk form. 
In addition, \easyedit ~provides a boolean switch, enabling users to either retain the pre-edit weights for single-instance editing or discard them for sequential editing.

\subsection{Method}
\label{sec:method}
As the core component of knowledge editing, editing methods alter the model's behavior by modifying its internal parameters (e.g. MLP, Attention Mechanisms) or explicitly utilizing preceding editing facts, among other strategies. Impressive related works (Table \ref{tab:edit_classifier}) abound in this field, and they can be generally grouped into three categories as proposed by \citet{yao2023editing}.

\paragraph{Memory-based}
This category, encompassing methods such as SERAC \cite{mitchell2022memorybased}, IKE \cite{zheng2023edit}, and GRACE \cite{hartvigsen2023aging}, emphasizes the use of memory elements to store and manipulate information during editing.
SERAC applies retrieval and classification routing, GRACE replaces hidden states with parameters searched from a codebook for edit memorization, while IKE uses context-edit facts to guide the model in generating edited facts.

\paragraph{Meta-learning}
These methods learn the weight updates (denoted as $\Delta$), which are then added to the original weights for editing.
Examples include KE \cite{decao2021editing}, which uses a bidirectional-LSTM to predict weight updates, and MEND \cite{mitchell2022fast}, which adjusts model parameters through low-rank decomposition of gradients.

\paragraph{Locate-Then-Edit}
This paradigm focuses on knowledge localization to modify the parameters of specific neurons responsible for storing the editing facts.
\easyedit ~integrates methods like KN \cite{DBLP:journals/corr/abs-2104-08696}, which employs gradient-based methods to update specific neurons.
Moreover, \easyedit ~supports ROME \cite{meng2023locating}, PMET \cite{li2024pmet} and MEMIT \cite{meng2022massediting}, leveraging causal intervention to pinpoint knowledge within a specific MLP layer and enabling the modification of the entire matrix.

However, it is not practical to expose the editing methods directly to users due to the complexity of the underlying concepts and the time investment required to understand them.
Additionally, differences in input-output formats across methods could further complicate the learning process.
To circumvent these hurdles, we implement a unified interface, \texttt{apply\_to\_model}, in \easyedit.
Aligning with the \emph{Strategy} design pattern, this interface is designed to be overridden by different types of editing methods, ensuring consistent input and output types.
Specifically, it accepts a `request' that includes the editing descriptor, the target of the edit, and any input data necessary to evaluate the editing performance. 
After processing the request(s), the interface returns the edited model weights.
This design ensures both flexibility and easy-to-use, enabling users to handle knowledge editing instances effortlessly and utilize the customized models in other downstream tasks.

\subsection{Hparams}

When initializing an editing method, it is crucial to specify the related hyperparameters.
These include the model to be edited, the layers targeted for modification, and, optionally, the type of external model, among other parameters.
For methods that alter the LLMs' internal parameters, the adjustable parameter names should be indicated using the \textsc{module\_name} format, such as \emph{transformer.h.5.mlp.fc\_out}.
In this case, the parameters of the \emph{fc\_out} linear layer in the fifth layer MLP of GPT-J would be modified, while all other parameters remain frozen.
Layer selection adheres to the locality of knowledge \cite{meng2023locating} or retains layers with higher success rates in pilot experiments \cite{mitchell2022fast}, as elaborated in Appendix \ref{sec:appendix_hparams}.

\begin{figure}[t]
    \includegraphics[width=0.48 \textwidth]
    {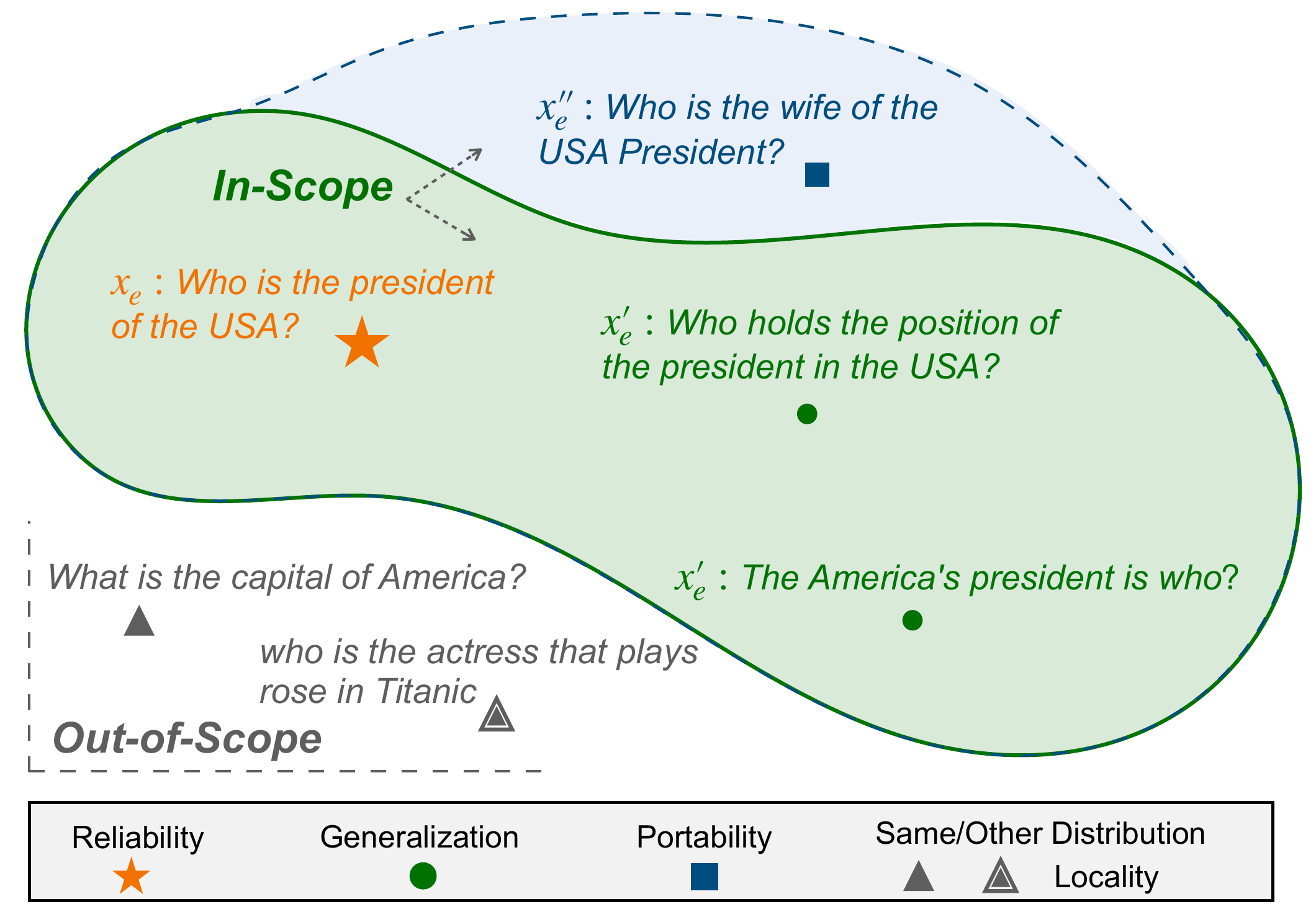}
    \caption{Depiction of the edit scope for edit descriptor \emph{Who is the president of the USA?} It contains an example for knowledge editing evaluation, including Reliability, Generalization, Portability, and Locality. 
    }
    \vspace*{-0.10in}
    \label{fig:metric-example}
\end{figure}

All hyperparameter classes derive from a common base class, \texttt{Hyperparams}, which includes necessary attributes and abstract methods.
This base class supports loading hyperparameters in both \emph{yaml} and \emph{json} formats. 
Moreover, the \texttt{Hyperparams} base class can be used to initialize the Trainer module, streamlining the workflow.

\subsection{Trainer}
\label{sec:trainer}

Certain editing methods, which employ meta-learning or utilize classifiers (as shown in Table \ref{tab:edit_classifier}), necessitate the training of additional parameters or the implementation of extra network structures.
Similar to Hyperparameters (\texttt{Hparams}), all Trainer classes inherit from a common base class, \texttt{BaseTrainer}. 
It includes essential attributes and abstract methods such as run and validate steps. 
Subclasses of the \texttt{BaseTrainer} define specific training steps for editing, such as calculating editing loss and locality loss, as well as the strategies for combining these losses.
Once additional network structures are obtained, the subsequent editing process follows the same path as the Training-Free method.
In \easyedit, various Trainers can be easily called with one click.

\section{Evaluation}
\label{sec:evaluate}

Knowledge editing, as defined by \citet{mitchell2022memorybased}, involves supplying a specific editing descriptor $x_e$ (input instance) and an editing target $y_e$ (desired output).
From these, an editing instance $z_e$ is generated in the form: $z_e \sim [x_e, y_e]$.
The goal is to adjust the behavior of the initial base model $f_{\theta}$ (where $\theta$ represents the model's parameters) to produce an edited model $f_{\theta_e}$.
Ideally, for the editing instance, the edited model would behave such that $f_{\theta_e}(x_e) = y_e$.
Additionally, the editing scope $S(z_e)$ refers to a set of input examples whose true labels have been influenced by the editing instance.
In most cases, a successful edit should affect the model's predictions for numerous In-Scope ($I(x_e) \sim \{x_e'|x_e' \in S(z_e)\}$) inputs, while leaving Out-of-Scope ($O(x_e) \sim \{x_e'|x_e' \notin S(z_e)\}$) inputs unchanged.

We employ six dimensions of metrics to assess the performance of editing methods, including \textbf{Reliability}, \textbf{Generalization}, \textbf{Locality}, \textbf{Portability}, \textbf{Fluency} \cite{zhang2018generating} and \textbf{Efficiency} (as shown in Figure \ref{fig:metric-example}).

\paragraph{Reliability}This metric measures the average accuracy on the given editing instance $z_e$.
\vspace{-0.05in}

\paragraph{Generalization} The edit should appropriately influence in-scope inputs, this metric gauges the average accuracy on in-scope inputs $I(x_e)$.
\vspace{-0.05in}

\paragraph{Locality} Editing should adhere to the principle of locality, it evaluates whether out-of-scope inputs $O(x_e)$ can remain unchanged as the base model.
\vspace{-0.05in}

\paragraph{Portability} The robust generalization of the edit, assessing whether the edited knowledge can be effectively applied to related content.
\vspace{-0.05in}

\paragraph{Fluency} It measures the weighted average of bi-gram and tri-gram entropies to assess the diversity of text generations.
\vspace{-0.05in}

\paragraph{Efficiency} Editing should be time and resource-efficient. 
This metric quantifies efficiency by measuring editing time and VRAM consumption.

\section{Experiments}

In this section, we will outline the experiment setting and report the empirical results of multiple editing methods supported in \easyedit ~(Table \ref{tab:result}).

\subsection{Experiment Setting}

To validate the potential application of knowledge editing on LLMs, we utilize \textbf{LlaMA 2} (7B) \cite{touvron2023llama2}, a model with a large parameter size, representing the decoder-only structure.

We employ the ZsRE dataset to test the capability of knowledge editing in incorporating substantial and general fact associations into the model.
ZsRE \cite{levy-etal-2017-zero} is a question-answering (QA) dataset that generates an equivalence neighbor through back-translation.
Later, it is further expanded by \citet{yao2023editing} to provide a more comprehensive evaluation of knowledge editing, including an assessment of the LLMs' ability to integrate the edited fact with other facts related to the target object o* (an aspect of Portability).
For baselines, we compare various editing methods and additionally employ FT-L from ROME \cite{meng2023locating}. 
FT-L updates parameters for a single MLP layer and applies an $L\infty$ norm constraint to limit the weight changes.

\subsection{Experiment Results}

Table \ref{tab:result} reveals SERAC and IKE's superior performance on the ZsRE datasets, exceeding 99\% on several metrics.
While ROME and MEMIT perform sub-optimally in generalization, they exhibit relatively high performance in terms of reliability and locality.
IKE exhibits the potential of gradient-free updates through in-context learning, leading to near-perfect scores in both reliability and generalization.
However, it shows some deficiency in locality, as preceding prompts may influence out-of-scope inputs.
GRACE exhibits poor generalization, possibly attributed to the lack of explicit semantic representation in its activations within the decoder-only model \cite{liu2023meaning}.
FT-L's performance on ZsRE falls significantly short compared to ROME, even though both methods modify the same layer parameters.
This suggests that under the norm constraint, fine-tuning is not an effective strategy for knowledge editing.
MEND performs well overall, achieving over 90\% accuracy on multiple metrics and even surpassing ROME in terms of reliability and generalization.
KN performs poorly, indicating that it may be better suited for editing tasks in smaller models or tasks involving knowledge attribution.

\definecolor{Mycolor1}{HTML}{BAD8F2}
\definecolor{Mycolor2}{HTML}{DDEEFA}

\newcommand{\first}{\colorbox{Mycolor1}}
\newcommand{\second}{\colorbox{Mycolor2}}

\begin{table}[t]
\centering
\small
\resizebox{1.0\columnwidth}{!}{
\begin{tabular}{lccccc}
\toprule
& \textbf{Reliability} &  \textbf{Generalization} &  \textbf{Locality} & \textbf{Portability} & \textbf{Fluency} \\
\midrule
FT-L          &56.94            &52.02                  & 96.32 & 51.03 & 488.41 \\
SERAC        &\second{99.49}            &\second{99.13}       & \second{100.00} &57.82 & 423.22 \\
IKE        &\first{\textbf{100.00}}            &\first{\textbf{99.98}}       &69.19 & \first{\textbf{67.56}} & 557.37 \\
MEND        &  94.24          &  90.27     & 97.04 &56.95 & 540.06 \\
KN        &28.95            &28.43      & 65.43 &37.18 & 478.32 \\
ROME        &92.45            &87.04    & 99.63 &\second{57.47}  & \first{\textbf{587.58}} \\
MEMIT       &92.94      &85.97  & 99.49 &60.64 & \second{576.51} \\
GRACE       &99.22      &0.43  & \first{\textbf{100.00}} &56.87 & 426.31 \\
\bottomrule
\end{tabular}
}
\caption{Editing results of the four metrics on LlaMA-2 using \easyedit. 
The settings for the model and the dataset are
the same with \citet{yao2023editing}.}
\vspace{-6mm}
\label{tab:result}
\end{table}

For the Portability evaluation, where the inference depends on a single connection or `hop' between facts, most editing methods struggle to effectively combine the edited fact with other facts relevant to the target object o*.
While SERAC obtains good performance on previous metrics, it completely fails to propagate the edited knowledge.
This is because SERAC utilizes an external model with a smaller parameter size for counterfactual routing whereas the smaller model struggles to recall a rich set of relevant facts. 
IKE still maintains a relatively high capability for ripple editing (exceeding 67\%), demonstrating that in-context learning is a promising approach to propagate edited knowledge to other related facts.


\section{Conclusion and Future work}
We propose \easyedit, an easy-to-use knowledge editing framework for LLMs, which supports many cutting-edge approaches and various LLMs.
The ability to edit and manipulate LLMs in a controlled and targeted manner may open up new possibilities for knowledge augmentation \cite{wu2023asdkb,wu2020knowledge, zhang2022knowledge, Chen_2022} and adaptation across various natural language processing tasks \cite{DBLP:journals/corr/abs-2307-10169}. 
In the future, we will continue to integrate advanced editing technologies 
into \easyedit, aiming at facilitating further research and inspiring new ideas for the NLP community.

\section*{Acknowledgments}
\label{sec:ack}
We thank the developers of the ROME\footnote{\url{https://github.com/kmeng01/rome}} library for their significant contributions to the NLP community.
We are grateful to Ting Lu and Yu Zhang who participated in the development of this project during the Zhejiang University Summer Camp.
We also extend our gratitude to the NLP team at East China Normal University, particularly Lang Yu, for their support of the Melo module.
Special thanks to Tom Hartvigsen for his contributions to the implementation of GRACE.
We are grateful to the TMG-NUDT team for their valuable suggestions and technical support for the PMET method. 
We are grateful to Jia-Chen Gu from the University of California, Los Angeles, and  Haiyang Yu from the Department of Cyberspace Security, University of Science for their constructive suggestions on development of {\easyedit}.
We thank Yiquan Wu and Zeqing Yuan for helping the AAAI 2024 tutorial (canceled since part of speakers cannot present in person) of EasyEdit. 
Appreciation is also extended to all PR contributors, and issue feedback providers during the EasyEdit version iterations, especially Damien de Mijolla for proposing different optimization goals for FT, which complemented the fine-tuning baseline, and to Yuxuan Zhai for pointing out the portability metric evaluation issue of LlaMA-2-7B.

We would like to express gratitude to the anonymous reviewers for their kind comments. 
This work was supported by the National Natural Science Foundation of China (No. 62206246, No. NSFCU23B2055, No. NSFCU19B2027), the Fundamental Research Funds for the Central Universities (226-2023-00138), Zhejiang Provincial Natural Science Foundation of China (No. LGG22F030011), Yongjiang Talent Introduction Programme (2021A-156-G), CCF-Tencent Rhino-Bird Open Research Fund, Tencent AI Lab Rhino-Bird Focused Research Program (RBFR2024003), Information Technology Center and State Key Lab of CAD\&CG, Zhejiang University.
 
\section*{Ethics Statement}
The significance of knowledge editing lies in its direct impact on the behavior and output results of LMs. Malicious edits may lead to the generation of responses with toxicity or bias in LMs, posing potential harm to users and society. Therefore, when applying knowledge editing techniques or utilizing this system, careful consideration must be given to potential risks and ethical concerns. All our data undergoes meticulous manual inspection, and any malicious edits or offensive content have been removed.

\bibliography{anthology, custom}
\bibliographystyle{acl_natbib}

\appendix

\section{Preliminaries of Model Editing}
\label{sec:editing_definition}

The task of knowledge editing is to effectively modify the initial base model $f_{\theta}$ to the edited model $f_{\theta'}$, with corresponding parameter adjustments for a specific input-output pair $(x_e,y_e)$, where $x_e \in \mathcal{X}_e$ and $f_{\theta}(x_e) \neq y_e$. Here, $\mathcal{X}_e$ represents the entire set to be edited. Therefore, the current problem formulation for knowledge editing can be broadly categorized into three types:

1.	\textbf{Single Instance Editing}: Evaluating the performance of the model after a single edit. The model reloads the original weights after a single edit:

\begin{equation}
    \theta' \leftarrow \mathop{\arg}\limits_{\theta} \min (\Vert f_\theta(x_e) - y_e \Vert)
\end{equation}

2.	\textbf{Batch Instance Editing}: Simultaneously modifying $N$ knowledge instances (where $N \ll \vert \mathcal{X}_e \vert$) and evaluating the performance of the edited model after processing a batch. The model reloads the original weights after processing a batch of edits:

\begin{equation}
    \theta' \leftarrow \mathop{\arg}\limits_{\theta} \min \sum\limits_{e=1}^{N} (\Vert f_\theta(x_e) - y_e \Vert)
\end{equation}

3.	\textbf{Sequential Editing}: This approach requires sequentially editing each knowledge instance, and evaluation must be performed after all knowledge updates have been applied:

\begin{equation}
    \theta' \leftarrow \mathop{\arg}\limits_{\theta} \min \sum\limits_{e=1}^{\vert \mathcal{X}_e \vert} (\Vert f_\theta(x_e) - y_e \Vert)
\end{equation}

\section{Default Hparams Settings}
\label{sec:appendix_hparams}

\textsc{EasyEdit} provides optimal hyperparameters for various editing methods.
In addition to common parameters such as learning rate, steps, and regularization coefficients, the location of layers for editing can also be considered as hyperparameters, significantly influencing the robustness of the editing process. The following tables demonstrate the default location settings in \textsc{EasyEdit} (using \textbf{Llama-2-7B} as an example).

\paragraph{ROME}
We follow \citet{meng2023locating} in utilizing causal mediation analysis to identify an intermediate layer in the model responsible for recalling facts. The causal traces reveal an early site (5th layer) with causal states concentrated at the last token of the subject, indicating a significant role for MLP states at that specific layer (Table \ref{tab:rome_layer}). 

\begin{table}[t]
\centering
\begin{tabular}{l} 
\toprule
Layer with Value Loss\\
\midrule
\small{\texttt{model.layers.31}}\\
\midrule
Target Layer for Updating Weights\\
\midrule
\small{\texttt{model.layers.5.mlp.down\_proj}}\\
\bottomrule 
\end{tabular}
\caption{Default Target Modules in \textbf{ROME}} 
\label{tab:rome_layer}
\end{table}

\begin{table}[t]
\centering
\begin{tabular}{l} 
\toprule
Layer with Value Loss\\
\midrule
\small{\texttt{model.layers.31}}\\
\midrule
Target Layer for Updating Weights\\
\midrule
\small{\texttt{model.layers.4.mlp.down\_proj}}\\
\small{\texttt{model.layers.5.mlp.down\_proj}}\\
\small{\texttt{model.layers.6.mlp.down\_proj}}\\
\small{\texttt{model.layers.7.mlp.down\_proj}}\\
\small{\texttt{model.layers.8.mlp.down\_proj}}\\
\bottomrule 
\end{tabular}
\caption{Default Target Modules in \textbf{MEMIT} and \textbf{PMET}} 
\label{tab:memit_layer}
\end{table}

\begin{table}[t]
\centering
\begin{tabular}{l} 
\toprule
CodeBook Target Modules\\
\midrule
\small{\texttt{model.layers[27].mlp.down\_proj.weight}}\\
\bottomrule 
\end{tabular}
\caption{Default Target Modules in \textbf{GRACE}} 
\label{tab:grace_layer}
\end{table}

\paragraph{MEMIT}
Following \citet{meng2022massediting}, we quantify the average indirect causal effect of MLP modules. 
The results demonstrate a concentration of intermediate states in LLaMA. The disparity in the effects between MLP severed and hidden states severed becomes significantly reduced after the 8th layer. 
We choose the entire critical range of MLP layers, denoted as $\mathcal{R} = \{4,5,6,7,8\}$ (Table \ref{tab:memit_layer}).

\paragraph{PMET}
PMET \cite{li2024pmet} adopts the localization strategy from MEMIT, designating the corresponding layer as the modification target. Building upon the update of MLP weights, PMET focuses on multi-head self-attention (MHSA), further substantiating the discovery that MHSA encodes specific patterns for general knowledge extraction. (Table \ref{tab:memit_layer}).

\paragraph{MEND}
In the context of meta-learning for editing, it is commonly observed that editing MLP layers yields better performance than editing attention layers.
Typically, MLP weights of the last 3 transformer blocks (totaling 6 weight matrices) are chosen for editing \cite{mitchell2022fast}. \textsc{EasyEdit} adheres to this default configuration (Table \ref{tab:mend_layer}).

\paragraph{GRACE}
Recent studies have revealed the impact of selecting the right layers for fine-tuning \cite{lee2023surgical}. Similarly, in GRACE \cite{hartvigsen2023aging}, we conduct pilot experiments, retaining layers with consistently high edit success rates (Table \ref{tab:grace_layer}).

\begin{table}[t]
\centering
\begin{tabular}{l} 
\toprule
Target Layer for Updating Weights\\
\midrule
\small{\texttt{model.layers.29.mlp.gate\_proj.weight}}\\
\small{\texttt{model.layers.29.mlp.up\_proj.weight}}\\
\small{\texttt{model.layers.29.mlp.down\_proj.weight}}\\
\small{\texttt{model.layers.30.mlp.gate\_proj.weight}}\\
\small{\texttt{model.layers.30.mlp.up\_proj.weight}}\\
\small{\texttt{model.layers.30.mlp.down\_proj.weight}}\\
\small{\texttt{model.layers.31.mlp.gate\_proj.weight}}\\
\small{\texttt{model.layers.31.mlp.up\_proj.weight}}\\
\small{\texttt{model.layers.31.mlp.down\_proj.weight}}\\
\bottomrule 
\end{tabular}
\caption{Default Target Modules in \textbf{MEND}} 
\label{tab:mend_layer}
\end{table}

\end{document}